\pdfoutput=1
\documentclass{article}
\usepackage{iclr2027_conference,times}
\usepackage{float}
\usepackage{amsmath,amsfonts,bm}

\def\eqref#1{equation~\ref{#1}}

\def\1{\bm{1}}

\DeclareMathAlphabet{\mathsfit}{\encodingdefault}{\sfdefault}{m}{sl}
\SetMathAlphabet{\mathsfit}{bold}{\encodingdefault}{\sfdefault}{bx}{n}

\usepackage{hyperref}
\hypersetup{hidelinks,
  pdftitle={AtomSearch: Evidence-Grounded Tree Search for Agent Skill Optimization},
  pdfauthor={Jia Tong, Senyou Deng, Min Peng, Jianfeng Pan}}
\usepackage{url}
\usepackage{graphicx}
\usepackage{algorithm}
\usepackage{algorithmic}
\usepackage{booktabs}
\usepackage{needspace}
\newlength{\pairedfigureheight}
\title{\centering AtomSearch: Evidence-Grounded Tree Search for Agent Skill Optimization}

\author{\hspace*{-\tabcolsep}\parbox{\textwidth}{\centering\vspace*{6pt}
{\large\bfseries Jia Tong, Senyou Deng, Min Peng, Jianfeng Pan}\\[8pt]
{\normalfont\large 360 Security Technology Inc. Beijing, China}\\[3pt]
{\small\textbf{Emails:} \normalfont tongjia1@360.cn, dengsenyou@360.cn, pengmin1@360.cn}}\hspace*{-\tabcolsep}}

\iclrfinalcopy
\begin{document}

\maketitle
\pagestyle{plain}

\begin{abstract}
Agent skills adapt frozen models through editable external instructions. Revising several rules at once makes changes difficult to inspect, while judging each local edit only by its immediate score can discard useful intermediate states. We introduce AtomSearch, which keeps revision actions local while using subsequent skill states to guide search. Contrastive atomic editing links each primitive rule edit to failed and successful executions, a behavioral hypothesis, and an intended scope. Skill-State Tree Search (STS) evaluates short edit compositions and selectively retains them as executable states that can be revisited and extended. Outcome memory supplies later proposals with observed repair and regression statistics. We evaluate AtomSearch across four tasks and five target models. With deepseek-v4-flash, it exceeds SkillOpt on three of four tasks, including a 9.68-percentage-point gain on LiveMathematicianBench, and SkillOpt's final skills contain 2.3--19.7 times as many characters. In 67\% of completed searches, the best validation state passes through an ancestor that was not highest-scoring at its depth. These results show how local rule edits can support the exploration of useful revision paths and yield compact skills, with additional offline evaluations used to explore alternative states.
\end{abstract}

\section{Introduction}
\label{sec:introduction}

Agent skills package task-specific instructions, procedures, scripts, and supporting resources into reusable artifacts that an agent can load at execution time~\citep{zhang2025agentskills}. Because the skill sits outside the model parameters, it can be revised without retraining the underlying model. Reusable artifacts also include executable code skills, as in Voyager~\citep{wang2023voyager}. We study how execution feedback can improve a fixed agent by revising its skill instructions.

The hard part is deciding what a revision should be and how its value should be judged. A candidate update may change several behavioral rules at once. If the score rises, the gain is assigned to the whole update rather than to a specific change; if it falls, the source of the regression is equally unclear. Repeating this process can also leave behind rules whose individual value was never established, so the skill grows even when some additions are redundant.

Recent work already addresses parts of this problem. SkillOpt restricts revisions to bounded add, delete, or replace operations and validates candidates on held-out data~\citep{yang2026skillopt}. Skill-$\alpha$ decomposes skill generation into individually evaluable sequential edits~\citep{shen2026skilla}, and SkillRevise uses execution traces to diagnose defects before applying targeted repairs~\citep{liu2026skillrevise}. Tree-based methods address a different limitation: bilevel skill optimization uses MCTS to search skill structure~\citep{huang2026bilevel}, while SkillHEX maintains persistent revision branches under sparse test-time feedback~\citep{feng2026skillhex}. These methods make revisions more controlled or broaden the search, but the revision unit and the horizon used to value that revision are still largely separate design choices.

Fine-grained edits can still interact. One edit may establish a condition under which a later edit becomes useful despite little immediate gain. It may also enable a correction that would not have been proposed from the original state. An optimizer that keeps only locally strong states can therefore discard useful paths. Figure~\ref{fig:search_tree} shows one such path, where a weaker intermediate state leads to the best state in the tree.

AtomSearch decouples the granularity of revision from the horizon over which revision paths are evaluated: each proposal specifies one primitive rule edit, while later outcomes guide exploration of its path. \textbf{Contrastive atomic editing} links the edit to failed and successful executions, a behavioral hypothesis, scope, expected effect, and operation rationale. The change and its evidence are traceable; execution measures the resulting skill's performance.

\textbf{Skill-State Tree Search (STS)} then evaluates that local action in a broader search context. Each node is a complete executable skill. Alternative revision paths remain searchable, and short deterministic rollouts test whether a primitive edit becomes useful after additional edits. When a rollout beats its primitive child, its composition becomes a persistent state that can be revisited. The edit stays local, but its value is allowed to emerge over later skill states.

Execution also reveals repair and regression outcomes that a scalar score hides. A revision can repair failed cases, break previously correct ones, or repeat a direction that has already been tried. We keep this information in a bounded \textbf{outcome memory} and expose selected records to later proposal rounds. The tree controls where search effort goes; the memory supplies the proposer with what earlier revisions actually did.

Repeated skill revision of this kind is closely related to recursive self-improvement (RSI), in which an agent keeps improving the artifacts that govern its own behavior~\citep{zelikman2023stop,zhang2025dgm,zhang2026ace,wu2026modularrsi}. The longer such a loop runs, the more it matters that each edit can be traced to its evidence and that earlier repairs and regressions are remembered; contrastive atomic editing and outcome memory provide both. The skill growth noted earlier matters as well, since every accepted change is another chance for the skill to lengthen.

\begin{figure}[t]
\centering
\includegraphics[width=\textwidth]{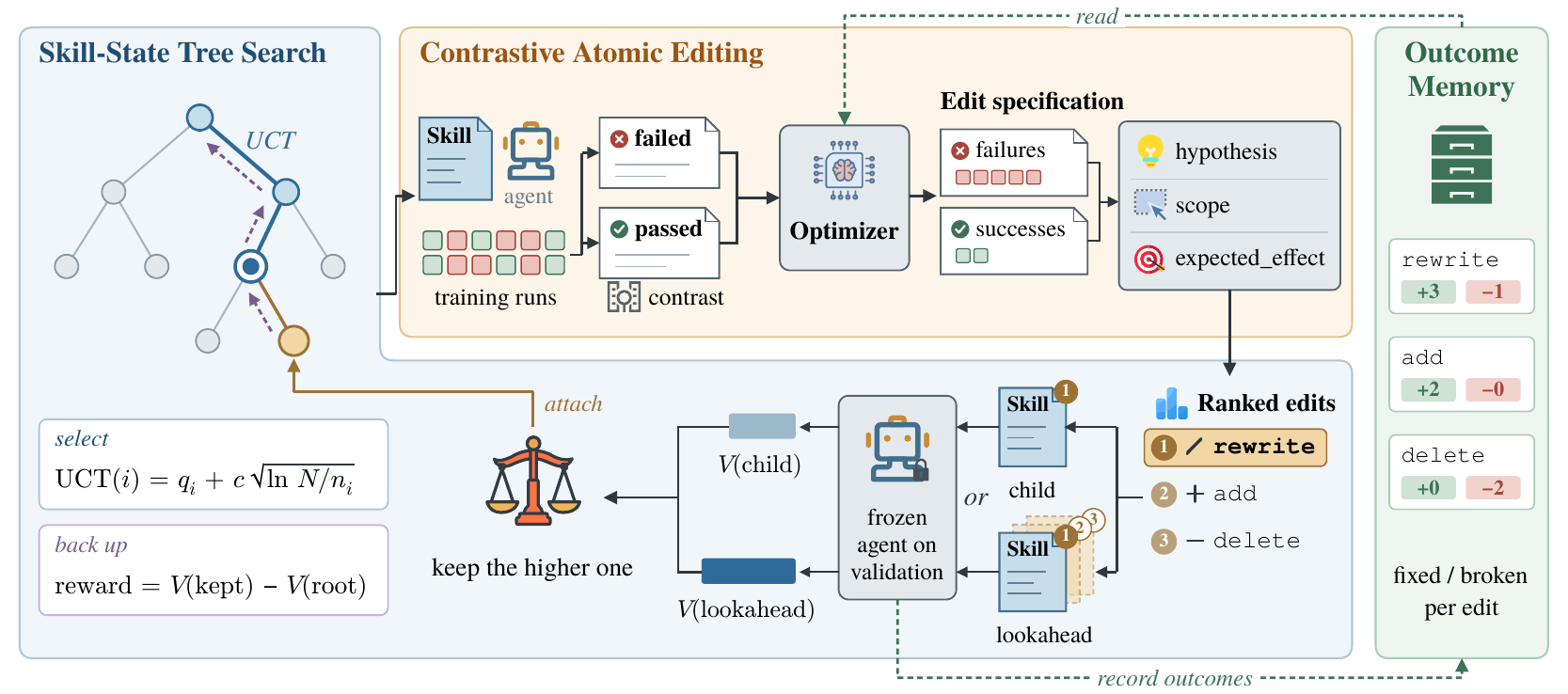}
\caption{Overview of AtomSearch. Contrastive atomic editing (top) contrasts failed and passed executions into edit specifications and ranked primitive edits. Skill-State Tree Search (left and bottom) applies the top-ranked edit at the node selected by UCT, also scores a short lookahead with up to two further edits, keeps the lookahead only if it beats the primitive child, and backs up the kept state's reward. Outcome memory (right) records repairs and regressions per edit for later proposals; its counts are schematic.}
\label{fig:overview}
\end{figure}

Our contributions are:
\begin{itemize}
    \item We define a contrastive atomic revision interface that binds execution evidence, a behavioral hypothesis, and scope to one primitive rule edit. Alternative corrections remain separate candidates that search can test and combine.
    \item We develop Skill-State Tree Search over complete executable skills. Lookahead compositions that improve on their primitive child are kept as tree nodes, allowing their outcomes to guide further exploration of earlier revision paths.
    \item We incorporate an outcome memory that supplies later proposals with attempted revision directions and repair/regression statistics. Results from a shared evaluation remain attached to the edit composition that produced them.
    \item Across four tasks and five targets, we measure performance, retained edit paths, skill size, and offline cost. With deepseek-v4-flash, AtomSearch exceeds SkillOpt on three tasks and produces shorter skills on all four; the recorded search trees also expose useful paths through locally weaker states.
\end{itemize}

\Needspace{7\baselineskip}
\section{Related Work}
\label{sec:related_work}

\paragraph{Localized and execution-grounded revision.}
Prompt optimizers search instructions~\citep{zhou2023ape,yang2024opro}, revise them using textual feedback~\citep{pryzant2023protegi,yuksekgonul2024textgrad}, or evolve candidate populations~\citep{guo2024evoprompt,fernando2024promptbreeder}. DSPy optimizes modular LM programs, and MIPRO jointly optimizes their instructions and demonstrations using end-to-end evaluation~\citep{khattab2024dspy,opsahlong2024mipro}. Other methods narrow the revision unit. ConstitutionalExperts edits individual principles~\citep{petridis2024constitutionalexperts}; AvaTaR contrasts successful and unsuccessful training examples to improve agent prompts~\citep{wu2024avatar}; Contrastive Reflection uses failures and nearby successful cases for targeted repairs~\citep{koh2026contrastive}. SkillOpt applies bounded edits under validation, Skill-$\alpha$ evaluates sequential edits, and SkillRevise diagnoses defects from execution traces~\citep{yang2026skillopt,shen2026skilla,liu2026skillrevise}. AtomSearch uses this feedback to specify one primitive rule edit as a search action, keeping its hypothesis and scope attached to the action.

\paragraph{Search over reasoning and reusable artifacts.}
Tree of Thoughts explores intermediate reasoning paths, while LATS combines tree search with actions, environment feedback, and reflection~\citep{yao2023tot,zhou2024lats}. These methods search task-solving trajectories. Search can also optimize artifacts reused across tasks: prompts, workflows, and heuristics~\citep{wang2024promptagent,zhang2025aflow,zheng2025mctsahd}. ADAS generates agent programs using an archive of previous designs~\citep{hu2025adas}. GEPA uses trajectory reflection to evolve prompts and maintains a Pareto set of complementary candidates~\citep{agrawal2026gepa}. For skills, bilevel optimization searches structure and refines content, while SkillHEX maintains revision branches under sparse feedback~\citep{huang2026bilevel,feng2026skillhex}. AtomSearch searches complete executable skills, with primitive edits as local actions; edit compositions that improve on their primitive child are kept as tree nodes.

\paragraph{Reusing feedback and maintaining skill content.}
Self-Refine and CRITIC use self-feedback and tool-based critique to revise outputs~\citep{madaan2023selfrefine,gou2024critic}. Across attempts or tasks, agents reuse verbal reflections and extracted experience~\citep{shinn2023reflexion,zhao2024expel}, induced workflows~\citep{wang2025awm}, and evolving strategy notes~\citep{suzgun2025dynamic}. ACE maintains a structured playbook through reflection and incremental updates~\citep{zhang2026ace}; SkillOpt also retains rejected-edit feedback and optimization guidance~\citep{yang2026skillopt}. Our memory stores revision directions and observed repair/regression statistics for later proposals, retaining joint provenance when edits share an evaluation. The amount of retained content is a separate design choice: ACE preserves detailed knowledge, whereas SkillReducer explicitly targets token efficiency~\citep{zhang2026ace,gao2026skillreducer}. AtomSearch has no length objective; its final skill sizes are measured outcomes.

\section{Method}
\label{sec:method}

Figure~\ref{fig:overview} summarizes the overall framework.
Given an initial skill $S_0$, a frozen target model $f$, and a held-out navigation set $\mathcal{D}_{\mathrm{nav}}$, we optimize only the skill text.
For a task-specific scoring function $m$, the value of a skill is

\begin{equation}
V(S)
=
\frac{1}{|\mathcal{D}_{\mathrm{nav}}|}
\sum_{(x,y)\in\mathcal{D}_{\mathrm{nav}}}
m\!\left(f(x;S),y\right),
\label{eq:skill_value}
\end{equation}
where $(x,y)$ is an input and its reference output in $\mathcal{D}_{\mathrm{nav}}$, and $f(x;S)$ is the output of $f$ on $x$ when $S$ is supplied as its skill.

The proposal unit is one primitive edit; the evaluated state may contain a composition of edits. Its measured score belongs to that state, while tree backup uses it to guide exploration of the preceding path.

\subsection{Contrastive Atomic Editing}
\label{sec:contrastive_editing}

We represent a skill as a sequence of addressable rule atoms. At each proposal round, the optimizer contrasts failed and successful executions and returns edit specifications together with their edits. A specification has five fields. \texttt{support\_ids} cites failed executions from the current minibatch that fail for the same underlying reason, and \texttt{counterexample\_ids} cites at least one successful execution whose behavior the edit must preserve. \texttt{hypothesis} states the behavioral distinction inferred from this contrast, \texttt{scope} states when it applies and when it does not, and \texttt{expected\_effect} names the failures to be repaired and the successful behavior to be kept. The optimizer also explains the chosen operation and edit count. Each edit cites one specification through \texttt{criterion\_id} and performs one primitive operation: \texttt{add}, \texttt{rewrite}, or \texttt{delete}; one specification justifies at most one edit.

Before search, a structural validator checks evidence-ID membership, minimum evidence counts, required fields, and the specification--edit mapping. It also removes duplicate edits and enforces the edit budget. These checks establish structural consistency, not the semantic correctness of the hypothesis; execution measures the resulting skill's observed performance.

Different corrections for the same failure pattern remain separate candidates. The search can therefore compare those directions without first merging them into a multi-rule patch.

\begin{figure}[t]
\centering
\includegraphics[width=0.96\linewidth]{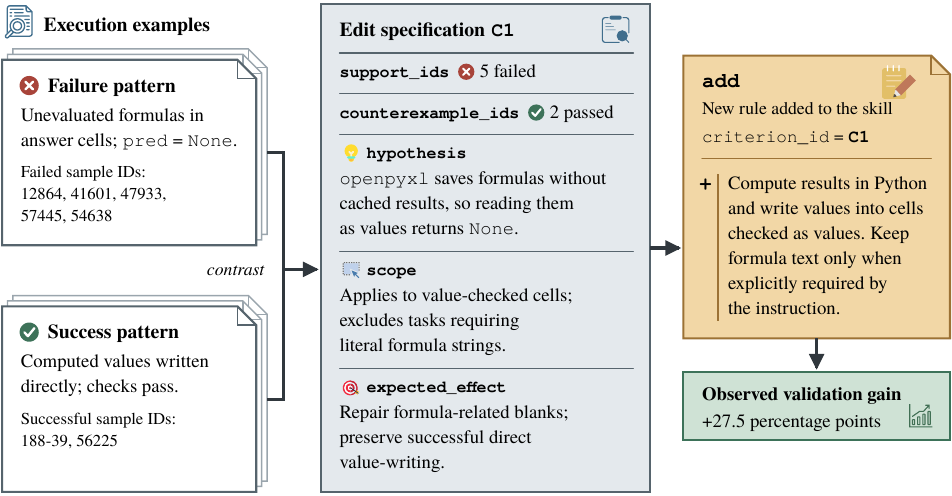}
\caption{A contrastive atomic edit from SpreadsheetBench. Failed and successful executions are contrasted into one edit specification, and the resulting \texttt{add} edit cites it through \texttt{criterion\_id}. Sample IDs are SpreadsheetBench task IDs. Pattern summaries and specification fields are condensed from the recorded proposal; the added rule is paraphrased. The observed gain is on the 40-item validation panel.}
\label{fig:edit-case}
\end{figure}

Figure~\ref{fig:edit-case} shows a concrete example: the proposed rule repairs formula-dependent value outputs while preserving tasks that explicitly require formula strings.

\subsection{Skill-State Tree Search}
\label{sec:sts}

A local edit can be easy to inspect and still be hard to value in isolation. An edit with little immediate gain may lead to a state from which a stronger correction becomes possible. We therefore search over complete skill states with \textbf{Skill-State Tree Search (STS)}, rooted at $S_0$. STS uses an MCTS backbone: each node stores an executable skill, and expansion starts from contrastively grounded primitive edits.

\paragraph{Selection and widening.}
At a node with skill state $S$, STS decides whether to expand $S$ once more or to descend into one of its existing children. The UCT selection rule~\citep{kocsis2006bandit,browne2012survey} scores both kinds of candidates: the current node itself, whose statistics come from the expansions made at $S$, and each child, whose statistics come from its subtree.
For a candidate $i$ with visit count $n_i$ and mean backed-up value $q_i$,

\begin{equation}
\operatorname{UCT}(i\mid S)
=
\begin{cases}
+\infty, & n_i=0, \\[4pt]
q_i
+
c_{\mathrm{uct}}
\sqrt{
\frac{\log \max\{N(S),1\}}
{n_i}
},
& n_i>0 ,
\end{cases}
\label{eq:uct}
\end{equation}

where $N(S)$ is the total number of visits through $S$ and $c_{\mathrm{uct}}=0.05$ is the exploration constant. For the current node, $n_i$ counts only the expansions made at $S$ itself.
The shared statistics place ``expand here'' and ``continue down an existing branch'' in the same budget-allocation decision.

To limit early branching, STS adopts progressive widening \citep{browne2012survey}.
In our implementation, the number of children currently eligible at $S$ is bounded by

\begin{equation}
K(S)
=
\max\!\left\{
1,\,
\min\!\left(
B,\,
\left\lceil
c_{\mathrm{pw}}
\max\{N(S),1\}^{\alpha}
\right\rceil
\right)
\right\},
\label{eq:pw}
\end{equation}

where $B=6$ is the maximum branch factor, $c_{\mathrm{pw}}=1.0$ controls the widening rate, and $\alpha=0.5$ is the widening exponent.
Here $K(S)$ only determines whether another child is eligible for exploration. It does not force an expansion. The current node still competes with its existing children under UCT for the next optimization attempt: progressive widening limits the available capacity, and UCT allocates it. Branch growth is therefore driven by the same observed returns that govern the rest of the tree, rather than by a fixed expand-as-soon-as-eligible schedule.

\paragraph{Expansion and execution-grounded lookahead.}
When the current node is selected for expansion, STS draws from the node's available contrastively grounded candidate edits, refilling this pool when necessary.
A selected primitive edit is applied to the current skill to produce a child state, which is evaluated on the navigation set.

The remaining ranked candidates from the same proposal pool provide a short lookahead. STS applies up to two of them in order, skipping conflicts, invalid edits, and no-ops. This rollout is deterministic. Only the final composed skill is executed and scored; intermediate rollout states are not evaluated separately.

\paragraph{Keeping compositions and backup.}
A composition is kept only if it scores strictly higher than the primitive child; ties keep the primitive child. Let $S'$ denote that primitive child and $S_{\mathrm{roll}}$ the final composed skill produced by the short rollout. STS adds to the tree
\begin{equation}
S_{\mathrm{keep}}=
\begin{cases}
S_{\mathrm{roll}}, & V(S_{\mathrm{roll}})>V(S'),\\
S', & \text{otherwise}.
\end{cases}
\label{eq:materialize}
\end{equation}
A composition that scores higher than $S'$ thus joins the tree as a single child node carrying all of its edits; otherwise only $S'$ is added. Intermediate rollout states are never added, and unused rollout edits remain available for later attempts.

The measured outcome belongs to the kept skill state, not independently to every edit in a kept composition. STS backs up only that state's value, using $V(S_{\mathrm{keep}})-V(S_0)$ as the reward. The same reward updates the expansion statistics of the node that acted and the subtree statistics of its ancestors, so later outcomes can affect which earlier revision paths receive more search.

At termination, STS uses a recursive adaptation of the robust-child recommendation rule \citep{browne2012survey}.
Starting from the root, it repeatedly selects the most visited child, stopping at a leaf or when the selected child has at most one visit.
The resulting state is returned, while the highest-value evaluated state is also retained. The revision-trajectory and tree analyses below use the latter to characterize discovered paths.

\subsection{Outcome Memory}
\label{sec:outcome_memory}

We supplement tree statistics with a bounded \textbf{outcome memory} of attempted revision directions and their repair/regression outcomes. Each record contains the attempted direction, an outcome tag, and the available aggregate repair and regression counts. When several edits share one evaluation, the record marks that result as belonging to the composition rather than assigning it to every edit separately. Failed attempts remain in the history as well.

Before another proposal round, the optimizer receives a bounded subset of these records: aggregate track records for explored directions together with selected past attempts. The memory is used only when proposing edits. It does not change UCT, progressive widening, or the search reward. We keep that boundary explicit because the two signals serve different purposes. Repair and regression counts summarize what a revision changed on individual examples, while tree values compare complete skill states on the navigation objective. Folding the former into UCT would mix those scales and make search allocation harder to interpret. The memory therefore changes what the proposer sees; execution still determines the value used by the tree.

The complete optimization procedure is given in Algorithm~\ref{alg:sts} in Appendix~\ref{app:algorithm}.

\section{Experiments}
\label{sec:experiments}

\subsection{Experimental Setup}
\label{sec:setup}

\paragraph{Tasks and models.}
We evaluate AtomSearch on Network Detection and Response (NDR), an in-house security task, and three public benchmarks: SpreadsheetBench~\citep{ma2024spreadsheetbench}, SearchQA~\citep{dunn2017searchqa}, and LiveMathematicianBench~\citep{he2026livemathematicianbench}. The public tasks cover spreadsheet manipulation, search-grounded question answering, and mathematical reasoning. The frozen target models are deepseek-v4-flash, GPT-5.2, GPT-5.5, Qwen3.5-4B, and Qwen3.6-35B-A3B. We optimize only the external skill.

\paragraph{NDR evaluation.}
NDR is an in-house dataset of HTTP traffic records constructed for network detection and response. All records are de-identified before use, and the dataset contains no personally identifiable information. For each record, the agent predicts an attack-status label (\texttt{IsAttack}), its type (\texttt{AttackType}), and its outcome (\texttt{AttackResult}). A prediction is correct only when the required fields jointly match the reference, including required empty outputs; attack types are compared as normalized sets. This setting tests whether explicit rules help distinguish evidence of an attack attempt from evidence that it succeeded. We evaluate the resulting skills offline on held-out records. NDR complements the public benchmarks with a domain-specific security task in which correct predictions require consistent reasoning across several labeled fields.

\paragraph{Comparison protocol.}
Our primary baseline is SkillOpt~\citep{yang2026skillopt}. The paired trajectory comparisons use the same underlying data splits and task-skill initialization. Execution feedback is used to construct candidate revisions, while a held-out validation panel (the navigation set $\mathcal{D}_{\mathrm{nav}}$) provides the signal used during optimization and skill selection. The test split is reserved for final reporting and is not used to generate revisions or guide tree search. For validation-trajectory analyses, both methods are evaluated on the same examples.

Edit count and search cost are different quantities in our analysis. Edit counts refer to textual rule edits on the retained trajectory; total target-model evaluations also include states explored and discarded by the tree.

\subsection{Performance Across Tasks and Target Models}
\label{sec:main_results}

Table~\ref{tab:main_results} reports test scores separately for each task and target. AtomSearch is higher on three of four tasks with both deepseek-v4-flash and GPT-5.2, on all four with Qwen3.6-35B-A3B, and on two with Qwen3.5-4B, with one tie. GPT-5.5 has two higher and two lower scores. These are point estimates, not significance claims.

On NDR, AtomSearch improves the reported score by 6.00 percentage points for both deepseek-v4-flash and GPT-5.2, and by 2.00 for Qwen3.6-35B-A3B. It ties Qwen3.5-4B and is 2.00 points lower with GPT-5.5. The NDR results therefore support offline applicability, with benefits that depend on the target model.

\begin{table}[t]
\centering
\caption{Test scores (\%) across target models and tasks. Bold marks the higher score within each pair, including ties.}
\label{tab:main_results}
\vspace{4pt}
\small
\setlength{\tabcolsep}{5pt}
\renewcommand{\arraystretch}{1.11}
\setlength{\aboverulesep}{0.55ex}
\setlength{\belowrulesep}{0.55ex}
\begin{tabular}{cccccc}
\toprule
Target model & Method & NDR & Spread. & SearchQA & LiveMath \\
\midrule
GPT-5.5 & SkillOpt & \textbf{73.00} & 77.50 & \textbf{87.10} & 61.30 \\
& AtomSearch & 71.00 & \textbf{79.64} & 85.57 & \textbf{65.32} \\
\midrule
deepseek-v4-flash & SkillOpt & 65.00 & 68.93 & \textbf{83.29} & 42.74 \\
& AtomSearch & \textbf{71.00} & \textbf{78.21} & 80.64 & \textbf{52.42} \\
\midrule
GPT-5.2 & SkillOpt & 66.00 & 57.10 & \textbf{83.10} & 36.00 \\
& AtomSearch & \textbf{72.00} & \textbf{57.50} & 82.50 & \textbf{41.13} \\
\midrule
Qwen3.5-4B & SkillOpt & \textbf{63.00} & 23.90 & 71.20 & \textbf{52.00} \\
& AtomSearch & \textbf{63.00} & \textbf{26.43} & \textbf{78.29} & 51.61 \\
\midrule
Qwen3.6-35B-A3B & SkillOpt & 66.00 & 47.50 & 80.30 & 41.60 \\
& AtomSearch & \textbf{68.00} & \textbf{59.64} & \textbf{80.79} & \textbf{43.55} \\
\bottomrule
\end{tabular}

\vspace{4pt}
\parbox{0.97\linewidth}{\footnotesize
Spread.: SpreadsheetBench; LiveMath: LiveMathematicianBench. SkillOpt scores with GPT-5.5 are its reported default-configuration results.}
\end{table}

\subsection{Validation Gains on Held-Out Test Data}
\label{sec:val-test}

We compare improvement over each run's initial skill on validation and test data. For a run, let $\Delta_{\mathrm{val}}=V_{\mathrm{final}}-V_{\mathrm{initial}}$ and $\Delta_{\mathrm{test}}=T_{\mathrm{final}}-T_{\mathrm{initial}}$. Table~\ref{tab:val-test} reports both gains and their ratio, $100\Delta_{\mathrm{test}}/\Delta_{\mathrm{val}}$, for six reported comparisons with positive validation gains.

AtomSearch has the larger ratio in all six comparisons; the median is 124\%, versus 82\% for SkillOpt. On SpreadsheetBench, both improve validation accuracy by 42.50 points, while their test gains are 38.21 and 30.00 points. Thus the observed validation improvements are accompanied by test improvements in these runs. The ratio compares gains, not final accuracy: on NDR with Qwen3.5-4B, both methods finish at 63.00 despite different ratios. Appendix~\ref{app:val-test} discusses these baselines and the scope of the comparison.

\begin{table}[t]
\centering\small
\caption{Validation and test gains over each run's recorded initial skill (percentage points). $R$ is the test gain divided by the validation gain, expressed as a percentage; it is not an accuracy. Ratios and medians are computed before rounding. Bold marks the larger $\Delta_{\mathrm{test}}$ and the larger $R$ within each pair.}
\label{tab:val-test}
\vspace{4pt}
\setlength{\tabcolsep}{4pt}
\renewcommand{\arraystretch}{1.10}
\setlength{\aboverulesep}{0.55ex}
\setlength{\belowrulesep}{0.55ex}
\begin{tabular}{@{}ccccccc@{}}
\toprule
 & \multicolumn{3}{c}{AtomSearch} & \multicolumn{3}{c}{SkillOpt} \\
\cmidrule(lr){2-4}\cmidrule(l){5-7}
Task / target & $\Delta_{\mathrm{val}}$ & $\Delta_{\mathrm{test}}$ & $R$ & $\Delta_{\mathrm{val}}$ & $\Delta_{\mathrm{test}}$ & $R$ \\
\midrule
Spreadsheet / deepseek & 42.50 & \textbf{38.21} & \textbf{90\%} & 42.50 & 30.00 & 71\% \\
NDR / deepseek & 11.43 & \textbf{15.00} & \textbf{131\%} & 13.33 & 12.00 & 90\% \\
NDR / GPT-5.2 & 20.00 & \textbf{27.00} & \textbf{135\%} & 20.95 & 20.00 & 95\% \\
NDR / GPT-5.5 & 24.13 & 20.00 & \textbf{83\%} & 29.52 & \textbf{24.00} & 81\% \\
NDR / Qwen3.5-4B & 12.38 & \textbf{20.00} & \textbf{162\%} & 18.10 & 15.00 & 83\% \\
NDR / Qwen3.6-35B-A3B & 8.57 & \textbf{10.00} & \textbf{117\%} & 15.24 & 8.00 & 52\% \\
\midrule
Median & & & \textbf{124\%} & & & 82\% \\
\bottomrule
\end{tabular}
\end{table}

\subsection{Revision Paths and Non-Myopic Search Behavior}
\label{sec:search_behavior}

\begin{figure}[t]
\centering
\includegraphics[width=\linewidth]{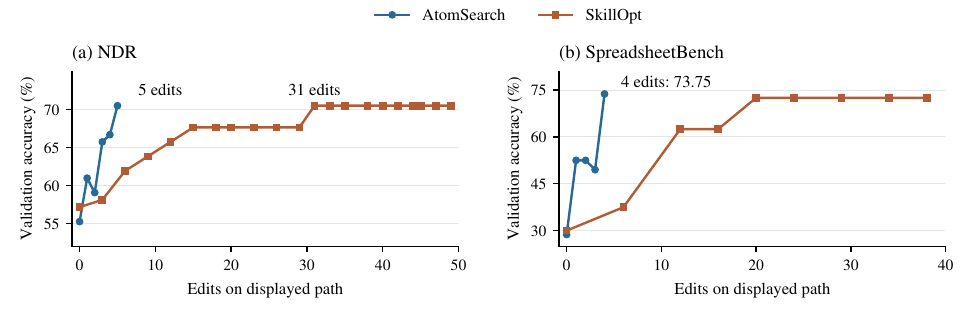}
\caption{Revision trajectories on shared validation panels. AtomSearch counts edits only along the displayed root-to-validation-argmax path, while SkillOpt follows its incumbent skill chain. All points are single recorded evaluations. Both paths include a score decrease before reaching their maxima. Edit counts exclude candidates explored off the displayed path.}
\label{fig:edit_efficiency}
\end{figure}

Final test scores hide the revision path. Figure~\ref{fig:edit_efficiency} instead plots validation accuracy against cumulative edits on the reported trajectory. AtomSearch follows the root-to-validation-argmax path and omits edits explored on other branches; SkillOpt follows its incumbent chain. These curves describe path length, not total search computation.

On NDR, AtomSearch reaches 70.48 after five retained edits; SkillOpt first reaches the same validation score after 31. On SpreadsheetBench, AtomSearch reaches 73.75 after four edits, while SkillOpt reaches 72.50 after 20 and stays there through edit 38. The tree uses additional evaluations to find those retained paths.

The SpreadsheetBench path is not monotonic. Validation accuracy is 52.50 after the second edit, falls to 49.50 after the third, and then rises to 73.75 after the fourth.

\begin{figure}[t]
\centering
\begin{minipage}[t]{0.49\textwidth}
\vspace{0pt}\centering
\includegraphics[height=\pairedfigureheight]{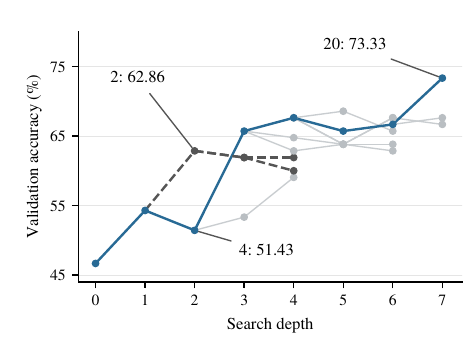}
\caption{An NDR search with GPT-5.2. The blue path passes through node~4 and reaches the best state, node~20. The dashed gray subtree starts at the locally stronger node~2 but never exceeds its score. Scores are validation accuracy.}
\label{fig:search_tree}
\end{minipage}\hfill
\begin{minipage}[t]{0.49\textwidth}
\vspace{0pt}\centering
\includegraphics[height=\pairedfigureheight]{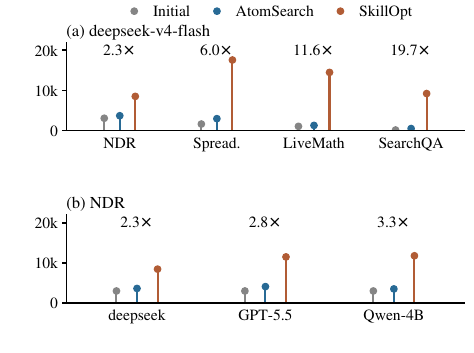}
\caption{Skill length in characters. (a) Four tasks with deepseek-v4-flash. (b) NDR across targets. Ratios divide SkillOpt length by AtomSearch length. Gray markers show the initial skill. Qwen-4B denotes Qwen3.5-4B.}
\label{fig:skill_length}
\end{minipage}
\end{figure}

Figure~\ref{fig:search_tree} shows the same pattern in an NDR search with GPT-5.2. Node~2 immediately improves from 54.29 to 62.86; node~4 is about 2.9 points below its parent. No descendant of node~2 exceeds node~2, whereas the node~4 branch later reaches node~20 at 73.33, the best validation score in the tree. Node~4 receives 23 visits and node~2 receives four.

In 20 of 30 completed searches (67\%), the best validation state has at least one ancestor that is not the highest-scoring state observed at the same depth. Such an ancestor need not be worse than its own parent, but a depth-wise immediate-score ranking would still prefer another branch.

\subsection{Compact Final Skills}
\label{sec:compactness}

Figure~\ref{fig:skill_length} compares the final artifacts with SkillOpt and with the initial skill. For deepseek-v4-flash, SkillOpt's final skills are 2.3--19.7 times as long across the four benchmarks. On NDR, the ratio is 2.3$\times$ to 3.3$\times$ across the three target models shown. AtomSearch does not optimize length, so the compact artifacts are an empirical outcome of the retained revision paths. In the ablation, the size drop appears as soon as contrastive atomic editing is added (16,536 to 2,352 characters; Table~\ref{tab:ablation}), so the reported task gains do not require continued rule accumulation.

\subsection{Component Analysis and Optimization Cost}
\label{sec:ablation_cost}

Table~\ref{tab:ablation} adds components cumulatively on SpreadsheetBench. Contrastive atomic editing accounts for most of the change: relative to Base, it raises the test score from 68.93 to 75.71 and shrinks the skill from 16,536 to 2,352 characters. Adding Skill-State Tree Search raises the score to 77.14, and adding outcome memory gives 78.21 for the full method. The later two steps are smaller (1.43 and 1.07 points, or 4 and 3 of 280 test items) and come from one task--target setting.

\begin{table}[t]
\centering
\caption{Cumulative ablation on SpreadsheetBench with deepseek-v4-flash. The first row is Base, the SkillOpt run reported in Table~\ref{tab:main_results}; each later row adds one component, and the last row is full AtomSearch, whose final skill is the one reported in Table~\ref{tab:main_results}. The initial skill has 1,594 characters. Bold marks the highest test score.}
\label{tab:ablation}
\vspace{4pt}
\small
\makeatletter\def\abcheck{\ensuremath{\mathchar"0\hexnumber@\symAMSa 58}}\makeatother
\setlength{\tabcolsep}{7pt}
\renewcommand{\arraystretch}{1.11}
\setlength{\aboverulesep}{0.55ex}
\setlength{\belowrulesep}{0.55ex}
\begin{tabular}{ccccc}
\toprule
\begin{tabular}[b]{@{}c@{}}Contrastive\\atomic editing\end{tabular} & \begin{tabular}[b]{@{}c@{}}Skill-State\\Tree Search\end{tabular} & \begin{tabular}[b]{@{}c@{}}Outcome\\memory\end{tabular} & Test (\%) & Characters \\
\midrule
-- & -- & -- & 68.93 & 16,536 \\
\abcheck & -- & -- & 75.71 & 2,352 \\
\abcheck & \abcheck & -- & 77.14 & 2,755 \\
\abcheck & \abcheck & \abcheck & \textbf{78.21} & 2,936 \\
\bottomrule
\end{tabular}
\end{table}

Short retained paths require additional offline evaluation. In the 17 documented runs, the optimizer is called 16--45 times; Appendix~\ref{app:cost} reports item-evaluation counts and cache usage.

\section{Discussion and Limitations}
\label{sec:discussion}

AtomSearch exchanges additional offline evaluation for exploration of alternative revision paths. This trade-off is most relevant when the resulting skill is reused across many executions. Adaptive rollout budgets and selective re-evaluation are natural next steps for reducing the search cost while preserving promising alternatives.

The separation between revision granularity and valuation horizon is deliberate. A small edit is easier to inspect, but its usefulness need not be visible at the same step. STS spends extra evaluations to preserve that distinction: the action remains local while later executable states can determine whether its path deserves more search. The traces in Section~\ref{sec:search_behavior} show that such paths occur in the current runs.

Finite validation panels and stochastic execution limit the precision of small score differences, and selecting among many candidates can overfit the validation criterion~\citep{cawley2010selectionbias}. The component ablation covers one task--target setting, where the increments from search and outcome memory are small; extending it across tasks and evaluation budgets would show how they vary. NDR is a single in-house task, and its results measure offline accuracy only.

\section{Conclusion}
\label{sec:conclusion}

AtomSearch keeps skill revisions local while allowing later edit combinations to inform search. Contrastive atomic editing makes each action and its evidence explicit; Skill-State Tree Search preserves evaluated compositions as extensible states; outcome memory carries observed revision outcomes into later proposals. Across the evaluated settings, AtomSearch produces compact skills and improves several task--model pairs. Its search traces show that locally weaker states can lead to the best discovered skill, at the cost of additional offline evaluation. Attributable edits and compact skills also carry over to RSI, where an agent that keeps revising its own skills needs both.

\label{page:main_end}
\clearpage
\bibliography{atomsearch_references}

\begin{thebibliography}{42}
\providecommand{\natexlab}[1]{#1}
\providecommand{\url}[1]{\texttt{#1}}
\expandafter\ifx\csname urlstyle\endcsname\relax
  \providecommand{\doi}[1]{doi: #1}\else
  \providecommand{\doi}{doi: \begingroup \urlstyle{rm}\Url}\fi

\bibitem[Agrawal et~al.(2026)Agrawal, Tan, Soylu, Ziems, Khare, Opsahl-Ong, Singhvi, Shandilya, Ryan, Jiang, Potts, Sen, Dimakis, Stoica, Klein, Zaharia, and Khattab]{agrawal2026gepa}
Lakshya~A. Agrawal, Shangyin Tan, Dilara Soylu, Noah Ziems, Rishi Khare, Krista Opsahl-Ong, Arnav Singhvi, Herumb Shandilya, Michael~J. Ryan, Meng Jiang, Christopher Potts, Koushik Sen, Alexandros~G. Dimakis, Ion Stoica, Dan Klein, Matei Zaharia, and Omar Khattab.
\newblock {GEPA}: Reflective prompt evolution can outperform reinforcement learning.
\newblock In \emph{International Conference on Learning Representations}, 2026.
\newblock URL \url{https://proceedings.iclr.cc/paper_files/paper/2026/hash/0e9e708b6f48e14fd0ac29e167413f76-Abstract-Conference.html}.

\bibitem[Browne et~al.(2012)Browne, Powley, Whitehouse, Lucas, Cowling, Rohlfshagen, Tavener, Perez, Samothrakis, and Colton]{browne2012survey}
Cameron~B. Browne, Edward Powley, Daniel Whitehouse, Simon~M. Lucas, Peter~I. Cowling, Philipp Rohlfshagen, Stephen Tavener, Diego Perez, Spyridon Samothrakis, and Simon Colton.
\newblock A survey of monte carlo tree search methods.
\newblock \emph{IEEE Transactions on Computational Intelligence and AI in Games}, 4\penalty0 (1):\penalty0 1--43, 2012.
\newblock \doi{10.1109/TCIAIG.2012.2186810}.

\bibitem[Cawley \& Talbot(2010)Cawley and Talbot]{cawley2010selectionbias}
Gavin~C. Cawley and Nicola L.~C. Talbot.
\newblock On over-fitting in model selection and subsequent selection bias in performance evaluation.
\newblock \emph{Journal of Machine Learning Research}, 11:\penalty0 2079--2107, 2010.
\newblock URL \url{https://jmlr.org/papers/v11/cawley10a.html}.

\bibitem[Dunn et~al.(2017)Dunn, Sagun, Higgins, Guney, Cirik, and Cho]{dunn2017searchqa}
Matthew Dunn, Levent Sagun, Mike Higgins, V.~Ugur Guney, Volkan Cirik, and Kyunghyun Cho.
\newblock {SearchQA}: A new q\&a dataset augmented with context from a search engine, 2017.
\newblock URL \url{https://arxiv.org/abs/1704.05179}.

\bibitem[Feng et~al.(2026)Feng, Chen, Zhao, Zhang, Wang, Lu, Wang, Xu, Li, and Chen]{feng2026skillhex}
Yuru Feng, Yaoqi Chen, Beidi Zhao, Qianxi Zhang, Xinjiang Wang, Jianan Lu, Zhirui Wang, Shusen Xu, Zengzhong Li, and Qi~Chen.
\newblock {SkillHEX}: Improving agent skills via hypothesis-driven autonomous exploration and exploitation, 2026.
\newblock URL \url{https://arxiv.org/abs/2608.05628}.

\bibitem[Fernando et~al.(2024)Fernando, Banarse, Michalewski, Osindero, and Rockt{\"a}schel]{fernando2024promptbreeder}
Chrisantha Fernando, Dylan~Sunil Banarse, Henryk Michalewski, Simon Osindero, and Tim Rockt{\"a}schel.
\newblock Promptbreeder: Self-referential self-improvement via prompt evolution.
\newblock In \emph{Proceedings of the 41st International Conference on Machine Learning}, volume 235 of \emph{Proceedings of Machine Learning Research}, pp.\  13481--13544. PMLR, 2024.
\newblock URL \url{https://proceedings.mlr.press/v235/fernando24a.html}.

\bibitem[Gao et~al.(2026)Gao, Li, Yuan, Ji, Ma, and Wang]{gao2026skillreducer}
Yudong Gao, Zongjie Li, Yuanyuan Yuan, Zimo Ji, Pingchuan Ma, and Shuai Wang.
\newblock {SkillReducer}: Optimizing {LLM} agent skills for token efficiency, 2026.
\newblock URL \url{https://arxiv.org/abs/2603.29919}.

\bibitem[Gou et~al.(2024)Gou, Shao, Gong, Shen, Yang, Duan, and Chen]{gou2024critic}
Zhibin Gou, Zhihong Shao, Yeyun Gong, Yelong Shen, Yujiu Yang, Nan Duan, and Weizhu Chen.
\newblock {CRITIC}: Large language models can self-correct with tool-interactive critiquing.
\newblock In \emph{International Conference on Learning Representations}, 2024.
\newblock URL \url{https://arxiv.org/abs/2305.11738}.

\bibitem[Guo et~al.(2024)Guo, Wang, Guo, Li, Song, Tan, Liu, Bian, and Yang]{guo2024evoprompt}
Qingyan Guo, Rui Wang, Junliang Guo, Bei Li, Kaitao Song, Xu~Tan, Guoqing Liu, Jiang Bian, and Yujiu Yang.
\newblock Connecting large language models with evolutionary algorithms yields powerful prompt optimizers.
\newblock In \emph{International Conference on Learning Representations}, 2024.
\newblock URL \url{https://arxiv.org/abs/2309.08532}.

\bibitem[He et~al.(2026)He, Yu, Dong, Liao, Xu, Goldblum, Bian, and Mesgarani]{he2026livemathematicianbench}
Linyang He, Qiyao Yu, Hanze Dong, Baohao Liao, Xinxing Xu, Micah Goldblum, Jiang Bian, and Nima Mesgarani.
\newblock {LiveMathematicianBench}: A live benchmark for mathematician-level reasoning with proof sketches, 2026.
\newblock URL \url{https://arxiv.org/abs/2604.01754}.

\bibitem[Hu et~al.(2025)Hu, Lu, and Clune]{hu2025adas}
Shengran Hu, Cong Lu, and Jeff Clune.
\newblock Automated design of agentic systems.
\newblock In \emph{International Conference on Learning Representations}, 2025.
\newblock URL \url{https://arxiv.org/abs/2408.08435}.

\bibitem[Huang et~al.(2026)Huang, Zhang, Xu, Zheng, and Lin]{huang2026bilevel}
Chenyi Huang, Haoting Zhang, Jingxu Xu, Zeyu Zheng, and Yunduan Lin.
\newblock Bilevel optimization of agent skills via monte carlo tree search, 2026.
\newblock URL \url{https://arxiv.org/abs/2604.15709}.

\bibitem[Khattab et~al.(2024)Khattab, Singhvi, Maheshwari, Zhang, Santhanam, Vardhamanan, Haq, Sharma, Joshi, Moazam, Miller, Zaharia, and Potts]{khattab2024dspy}
Omar Khattab, Arnav Singhvi, Paridhi Maheshwari, Zhiyuan Zhang, Keshav Santhanam, Sri Vardhamanan, Saiful Haq, Ashutosh Sharma, Thomas Joshi, Hanna Moazam, Heather Miller, Matei Zaharia, and Christopher Potts.
\newblock {DSPy}: Compiling declarative language model calls into state-of-the-art pipelines.
\newblock In \emph{International Conference on Learning Representations}, 2024.
\newblock URL \url{https://arxiv.org/abs/2310.03714}.

\bibitem[Kocsis \& Szepesv{\'a}ri(2006)Kocsis and Szepesv{\'a}ri]{kocsis2006bandit}
Levente Kocsis and Csaba Szepesv{\'a}ri.
\newblock Bandit based monte-carlo planning.
\newblock In \emph{Machine Learning: ECML 2006}, pp.\  282--293. Springer, 2006.

\bibitem[Koh et~al.(2026)Koh, Mo, Le, Zhan, Zheng, Bevis, Owen, Charney, Liu, and Wu]{koh2026contrastive}
Derek Koh, Jinghui Mo, Benjamin~H. Le, Jiening Zhan, Baofen Zheng, Kevin Bevis, Nathaniel~C. Owen, Lauren~Elizabeth Charney, Wenqiong Liu, and Jingwei Wu.
\newblock Contrastive reflection for iterative prompt optimization, 2026.
\newblock URL \url{https://arxiv.org/abs/2606.30840}.

\bibitem[Liu et~al.(2026)Liu, Su, Xie, Zhang, Zong, Guo, Xie, Ji, Yim, Luo, Ren, Ruan, Li, and Song]{liu2026skillrevise}
Yuxuan Liu, Zhaochen Su, Lingyun Xie, Yuhao Zhang, Qing Zong, Jiahe Guo, Zhongwei Xie, Yiyan Ji, Yauwai Yim, Hongyu Luo, Xiyu Ren, Chenyu Ruan, Haoran Li, and Yangqiu Song.
\newblock {SkillRevise}: Improving llm-authored agent skills via trace-conditioned skill revision, 2026.
\newblock URL \url{https://arxiv.org/abs/2606.01139}.

\bibitem[Ma et~al.(2024)Ma, Zhang, Zhang, Yu, Zhang, Zhang, Luo, Wang, and Tang]{ma2024spreadsheetbench}
Zeyao Ma, Bohan Zhang, Jing Zhang, Jifan Yu, Xiaokang Zhang, Xiaohan Zhang, Sijia Luo, Xi~Wang, and Jie Tang.
\newblock {SpreadsheetBench}: Towards challenging real world spreadsheet manipulation.
\newblock In \emph{Advances in Neural Information Processing Systems}, volume~37, 2024.
\newblock \doi{10.52202/079017-3007}.
\newblock URL \url{https://arxiv.org/abs/2406.14991}.

\bibitem[Madaan et~al.(2023)Madaan, Tandon, Gupta, Hallinan, Gao, Wiegreffe, Alon, Dziri, Prabhumoye, Yang, Gupta, Majumder, Hermann, Welleck, Yazdanbakhsh, and Clark]{madaan2023selfrefine}
Aman Madaan, Niket Tandon, Prakhar Gupta, Skyler Hallinan, Luyu Gao, Sarah Wiegreffe, Uri Alon, Nouha Dziri, Shrimai Prabhumoye, Yiming Yang, Shashank Gupta, Bodhisattwa~Prasad Majumder, Katherine Hermann, Sean Welleck, Amir Yazdanbakhsh, and Peter Clark.
\newblock Self-refine: Iterative refinement with self-feedback.
\newblock In \emph{Advances in Neural Information Processing Systems}, volume~36, 2023.
\newblock URL \url{https://arxiv.org/abs/2303.17651}.

\bibitem[Opsahl-Ong et~al.(2024)Opsahl-Ong, Ryan, Purtell, Broman, Potts, Zaharia, and Khattab]{opsahlong2024mipro}
Krista Opsahl-Ong, Michael~J. Ryan, Josh Purtell, David Broman, Christopher Potts, Matei Zaharia, and Omar Khattab.
\newblock Optimizing instructions and demonstrations for multi-stage language model programs.
\newblock In \emph{Proceedings of the 2024 Conference on Empirical Methods in Natural Language Processing}, pp.\  9340--9366, 2024.
\newblock \doi{10.18653/v1/2024.emnlp-main.525}.
\newblock URL \url{https://aclanthology.org/2024.emnlp-main.525/}.

\bibitem[Petridis et~al.(2024)Petridis, Wedin, Yuan, Wexler, and Thain]{petridis2024constitutionalexperts}
Savvas Petridis, Ben Wedin, Ann Yuan, James Wexler, and Nithum Thain.
\newblock {ConstitutionalExperts}: Training a mixture of principle-based prompts.
\newblock In \emph{Proceedings of the 62nd Annual Meeting of the Association for Computational Linguistics (Volume 2: Short Papers)}, pp.\  574--582, Bangkok, Thailand, 2024. Association for Computational Linguistics.
\newblock \doi{10.18653/v1/2024.acl-short.52}.
\newblock URL \url{https://aclanthology.org/2024.acl-short.52/}.

\bibitem[Pryzant et~al.(2023)Pryzant, Iter, Li, Lee, Zhu, and Zeng]{pryzant2023protegi}
Reid Pryzant, Dan Iter, Jerry Li, Yin Lee, Chenguang Zhu, and Michael Zeng.
\newblock Automatic prompt optimization with ``gradient descent'' and beam search.
\newblock In \emph{Proceedings of the 2023 Conference on Empirical Methods in Natural Language Processing}, pp.\  7957--7968, 2023.
\newblock \doi{10.18653/v1/2023.emnlp-main.494}.
\newblock URL \url{https://aclanthology.org/2023.emnlp-main.494/}.

\bibitem[Shen et~al.(2026)Shen, Zhang, Guo, and Cheng]{shen2026skilla}
Junhao Shen, Zhanqiu Zhang, Yiwen Guo, and Hong Cheng.
\newblock Progressive agent skill generation via reinforcement learning, 2026.
\newblock URL \url{https://arxiv.org/abs/2608.01678}.

\bibitem[Shinn et~al.(2023)Shinn, Cassano, Gopinath, Narasimhan, and Yao]{shinn2023reflexion}
Noah Shinn, Federico Cassano, Ashwin Gopinath, Karthik Narasimhan, and Shunyu Yao.
\newblock {Reflexion}: Language agents with verbal reinforcement learning.
\newblock In \emph{Advances in Neural Information Processing Systems}, volume~36, 2023.
\newblock URL \url{https://proceedings.neurips.cc/paper_files/paper/2023/hash/1b44b878bb782e6954cd888628510e90-Abstract-Conference.html}.

\bibitem[Suzgun et~al.(2025)Suzgun, Yuksekgonul, Bianchi, Jurafsky, and Zou]{suzgun2025dynamic}
Mirac Suzgun, Mert Yuksekgonul, Federico Bianchi, Dan Jurafsky, and James Zou.
\newblock Dynamic cheatsheet: Test-time learning with adaptive memory, 2025.
\newblock URL \url{https://arxiv.org/abs/2504.07952}.

\bibitem[Wang et~al.(2023)Wang, Xie, Jiang, Mandlekar, Xiao, Zhu, Fan, and Anandkumar]{wang2023voyager}
Guanzhi Wang, Yuqi Xie, Yunfan Jiang, Ajay Mandlekar, Chaowei Xiao, Yuke Zhu, Linxi Fan, and Anima Anandkumar.
\newblock Voyager: An open-ended embodied agent with large language models, 2023.
\newblock URL \url{https://arxiv.org/abs/2305.16291}.

\bibitem[Wang et~al.(2024)Wang, Li, Wang, Bai, Luo, Zhang, Jojic, Xing, and Hu]{wang2024promptagent}
Xinyuan Wang, Chenxi Li, Zhen Wang, Fan Bai, Haotian Luo, Jiayou Zhang, Nebojsa Jojic, Eric~P. Xing, and Zhiting Hu.
\newblock {PromptAgent}: Strategic planning with language models enables expert-level prompt optimization.
\newblock In \emph{International Conference on Learning Representations}, 2024.
\newblock URL \url{https://openreview.net/forum?id=22pyNMuIoa}.

\bibitem[Wang et~al.(2025)Wang, Mao, Fried, and Neubig]{wang2025awm}
Zora~Zhiruo Wang, Jiayuan Mao, Daniel Fried, and Graham Neubig.
\newblock Agent workflow memory.
\newblock In \emph{Proceedings of the 42nd International Conference on Machine Learning}, volume 267 of \emph{Proceedings of Machine Learning Research}, pp.\  63897--63911. PMLR, 2025.
\newblock URL \url{https://proceedings.mlr.press/v267/wang25bx.html}.

\bibitem[Wu et~al.(2024)Wu, Zhao, Huang, Huang, Yasunaga, Cao, Ioannidis, Subbian, Leskovec, and Zou]{wu2024avatar}
Shirley Wu, Shiyu Zhao, Qian Huang, Kexin Huang, Michihiro Yasunaga, Kaidi Cao, Vassilis~N. Ioannidis, Karthik Subbian, Jure Leskovec, and James Zou.
\newblock {AvaTaR}: Optimizing {LLM} agents for tool usage via contrastive reasoning.
\newblock In \emph{Advances in Neural Information Processing Systems}, volume~37, 2024.
\newblock URL \url{https://proceedings.neurips.cc/paper_files/paper/2024/hash/2db8ce969b000fe0b3fb172490c33ce8-Abstract-Conference.html}.

\bibitem[Wu et~al.(2026)Wu, Ren, Li, Li, Yang, Zhang, Gu, Yang, Batista-Navarro, Zhang, Liu, Zhou, Dai, and Lin]{wu2026modularrsi}
Siwei Wu, Jincheng Ren, Yizhi Li, Haau-Sing Li, Chengran Yang, Yuxuan Zhang, Weicheng Gu, Jian Yang, Riza Batista-Navarro, Chuanyi Zhang, Xianglong Liu, Ming Zhou, Bryan Dai, and Chenghua Lin.
\newblock {ModularRSI}: Modular and generalizable recursive harness self-improvement, 2026.
\newblock URL \url{https://arxiv.org/abs/2609.14857}.

\bibitem[Yang et~al.(2024)Yang, Wang, Lu, Liu, Le, Zhou, and Chen]{yang2024opro}
Chengrun Yang, Xuezhi Wang, Yifeng Lu, Hanxiao Liu, Quoc~V. Le, Denny Zhou, and Xinyun Chen.
\newblock Large language models as optimizers.
\newblock In \emph{International Conference on Learning Representations}, 2024.
\newblock URL \url{https://arxiv.org/abs/2309.03409}.

\bibitem[Yang et~al.(2026)Yang, Gong, Huang, Yang, Zhou, Huang, Li, Gao, Dai, Liu, Qiu, Yang, Chen, Yang, and Luo]{yang2026skillopt}
Yifan Yang, Ziyang Gong, Weiquan Huang, Qihao Yang, Ziwei Zhou, Zisu Huang, Yan Li, Xuemei Gao, Qi~Dai, Bei Liu, Kai Qiu, Yuqing Yang, Dongdong Chen, Xue Yang, and Chong Luo.
\newblock {SkillOpt}: Executive strategy for self-evolving agent skills.
\newblock In \emph{Advances in Neural Information Processing Systems}, 2026.
\newblock URL \url{https://arxiv.org/abs/2605.23904}.

\bibitem[Yao et~al.(2023)Yao, Yu, Zhao, Shafran, Griffiths, Cao, and Narasimhan]{yao2023tot}
Shunyu Yao, Dian Yu, Jeffrey Zhao, Izhak Shafran, Thomas~L. Griffiths, Yuan Cao, and Karthik Narasimhan.
\newblock Tree of thoughts: Deliberate problem solving with large language models.
\newblock In \emph{Advances in Neural Information Processing Systems}, volume~36, 2023.
\newblock URL \url{https://arxiv.org/abs/2305.10601}.

\bibitem[Yuksekgonul et~al.(2024)Yuksekgonul, Bianchi, Boen, Liu, Huang, Guestrin, and Zou]{yuksekgonul2024textgrad}
Mert Yuksekgonul, Federico Bianchi, Joseph Boen, Sheng Liu, Zhi Huang, Carlos Guestrin, and James Zou.
\newblock {TextGrad}: Automatic ``differentiation'' via text, 2024.
\newblock URL \url{https://arxiv.org/abs/2406.07496}.

\bibitem[Zelikman et~al.(2023)Zelikman, Lorch, Mackey, and Kalai]{zelikman2023stop}
Eric Zelikman, Eliana Lorch, Lester Mackey, and Adam~Tauman Kalai.
\newblock Self-taught optimizer ({STOP}): Recursively self-improving code generation, 2023.
\newblock URL \url{https://arxiv.org/abs/2310.02304}.

\bibitem[Zhang et~al.(2025{\natexlab{a}})Zhang, Lazuka, and Murag]{zhang2025agentskills}
Barry Zhang, Keith Lazuka, and Mahesh Murag.
\newblock Equipping agents for the real world with agent skills.
\newblock Anthropic Engineering, October 2025{\natexlab{a}}.
\newblock URL \url{https://www.anthropic.com/engineering/equipping-agents-for-the-real-world-with-agent-skills}.
\newblock Published October 16, 2025.

\bibitem[Zhang et~al.(2025{\natexlab{b}})Zhang, Hu, Lu, Lange, and Clune]{zhang2025dgm}
Jenny Zhang, Shengran Hu, Cong Lu, Robert Lange, and Jeff Clune.
\newblock Darwin {G}{\"o}del machine: Open-ended evolution of self-improving agents, 2025{\natexlab{b}}.
\newblock URL \url{https://arxiv.org/abs/2505.22954}.

\bibitem[Zhang et~al.(2025{\natexlab{c}})Zhang, Xiang, Yu, Teng, Chen, Chen, Zhuge, Cheng, Hong, Wang, Zheng, Liu, Luo, and Wu]{zhang2025aflow}
Jiayi Zhang, Jinyu Xiang, Zhaoyang Yu, Fengwei Teng, Xiong-Hui Chen, Jiaqi Chen, Mingchen Zhuge, Xin Cheng, Sirui Hong, Jinlin Wang, Bingnan Zheng, Bang Liu, Yuyu Luo, and Chenglin Wu.
\newblock {AFlow}: Automating agentic workflow generation.
\newblock In \emph{The Thirteenth International Conference on Learning Representations}, 2025{\natexlab{c}}.
\newblock URL \url{https://openreview.net/forum?id=z5uVAKwmjf}.

\bibitem[Zhang et~al.(2026)Zhang, Hu, Upasani, Ma, Hong, Kamanuru, Rainton, Wu, Ji, Li, Thakker, Zou, and Olukotun]{zhang2026ace}
Qizheng Zhang, Changran Hu, Shubhangi Upasani, Boyuan Ma, Fenglu Hong, Vamsidhar Kamanuru, Jay Rainton, Chen Wu, Mengmeng Ji, Hanchen Li, Urmish Thakker, James Zou, and Kunle Olukotun.
\newblock Agentic context engineering: Evolving contexts for self-improving language models.
\newblock In \emph{International Conference on Learning Representations}, 2026.
\newblock URL \url{https://arxiv.org/abs/2510.04618}.

\bibitem[Zhao et~al.(2024)Zhao, Huang, Xu, Lin, Liu, and Huang]{zhao2024expel}
Andrew Zhao, Daniel Huang, Quentin Xu, Matthieu Lin, Yong-Jin Liu, and Gao Huang.
\newblock {ExpeL}: {LLM} agents are experiential learners.
\newblock \emph{Proceedings of the AAAI Conference on Artificial Intelligence}, 38\penalty0 (17):\penalty0 19632--19642, 2024.
\newblock \doi{10.1609/aaai.v38i17.29936}.
\newblock URL \url{https://ojs.aaai.org/index.php/AAAI/article/view/29936}.

\bibitem[Zheng et~al.(2025)Zheng, Xie, Wang, and Hooi]{zheng2025mctsahd}
Zhi Zheng, Zhuoliang Xie, Zhenkun Wang, and Bryan Hooi.
\newblock Monte carlo tree search for comprehensive exploration in {LLM}-based automatic heuristic design.
\newblock In \emph{Proceedings of the 42nd International Conference on Machine Learning}, volume 267 of \emph{Proceedings of Machine Learning Research}, pp.\  78338--78373. PMLR, 2025.
\newblock URL \url{https://proceedings.mlr.press/v267/zheng25o.html}.

\bibitem[Zhou et~al.(2024)Zhou, Yan, Shlapentokh-Rothman, Wang, and Wang]{zhou2024lats}
Andy Zhou, Kai Yan, Michal Shlapentokh-Rothman, Haohan Wang, and Yu-Xiong Wang.
\newblock Language agent tree search unifies reasoning, acting, and planning in language models.
\newblock In \emph{Proceedings of the 41st International Conference on Machine Learning}, 2024.
\newblock URL \url{https://arxiv.org/abs/2310.04406}.

\bibitem[Zhou et~al.(2023)Zhou, Muresanu, Han, Paster, Pitis, Chan, and Ba]{zhou2023ape}
Yongchao Zhou, Andrei~Ioan Muresanu, Ziwen Han, Keiran Paster, Silviu Pitis, Harris Chan, and Jimmy Ba.
\newblock Large language models are human-level prompt engineers.
\newblock In \emph{International Conference on Learning Representations}, 2023.
\newblock URL \url{https://arxiv.org/abs/2211.01910}.

\end{thebibliography}
\bibliographystyle{iclr2027_conference}

\clearpage
\appendix
\section{Optimization Procedure}
\label{app:algorithm}

\begin{algorithm}[H]
\caption{Skill-State Tree Search with contrastive atomic editing and outcome memory}
\label{alg:sts}
\begin{algorithmic}[1]
\REQUIRE Initial skill $S_0$, search budget $T$
\STATE Initialize tree $\mathcal{T}$ with root $S_0$ and outcome memory $\mathcal{M}$
\FOR{$t=1,\ldots,T$}
    \STATE Select an expandable state $S$ using UCT under PW eligibility
    \STATE Obtain contrastively grounded candidate edits for $S$ using current evidence and $\mathcal{M}$ if needed
    \IF{no valid candidate remains}
        \STATE Back up the terminal outcome and \textbf{continue}
    \ENDIF
    \STATE Apply the highest-ranked untried primitive edit to obtain $S'$
    \STATE Evaluate $S'$ on $\mathcal{D}_{\mathrm{nav}}$
    \STATE Build and evaluate a short rollout from remaining ranked edits
    \IF{the rollout state scores higher than $S'$}
        \STATE $S_{\mathrm{keep}} \gets$ rollout state
    \ELSE
        \STATE $S_{\mathrm{keep}} \gets S'$
    \ENDIF
    \STATE Attach $S_{\mathrm{keep}}$ and back up $V(S_{\mathrm{keep}})-V(S_0)$
    \STATE Append the observed revision outcome(s) to $\mathcal{M}$
\ENDFOR
\STATE \textbf{return} the recursively selected robust child
\end{algorithmic}
\end{algorithm}

\section{Experimental details}
\label{app:details}

\subsection{Data splits and search configuration}
\label{app:setup}

Each benchmark is split once into a learn set (source of the failed and
successful executions contrasted by the proposer), a validation set (used
to accept edits and select the returned skill), and a held-out test set that
is read only for reporting. Table~\ref{tab:splits} lists the sizes.

\begin{table}[htbp]
\centering\small
\caption{Split sizes (items) and contrastive minibatch size per benchmark.
Minibatches use approximately one third of the learn set. LiveMath: LiveMathematicianBench.}
\label{tab:splits}
\vspace{4pt}
\begin{tabular}{@{}ccccc@{}}
\toprule
Benchmark & Learn & Validation & Test & Minibatch \\
\midrule
NDR              & 100 & 105 & 100  & 35  \\
SpreadsheetBench & 80  & 40  & 280  & 27  \\
SearchQA         & 400 & 200 & 1400 & 134 \\
LiveMath         & 35  & 18  & 124  & 12  \\
\bottomrule
\end{tabular}
\end{table}

The reference search configuration uses $k=5$ candidate specifications
per proposal round with an edit budget of 4, maximum branch factor $B=6$, maximum
depth 12, UCT exploration constant $c_{\mathrm{uct}}=0.05$, progressive widening with
$\alpha=0.5$ and $c_{\mathrm{pw}}=1.0$, and rollouts of at most two edits. Runs use 20 or 30 attempts (Table~\ref{tab:cost}). Target models are
queried with reasoning effort \emph{medium} and a completion limit of
16{,}384 tokens (24{,}576 for Qwen3.6-35B-A3B on NDR). Open-weight targets
are served locally with vLLM.

\subsection{Evaluation cache and optimization cost}
\label{app:cost}

Evaluation results are cached per (skill state, item, draw). Repeated reads
of an unchanged state are served from the cache and are not counted against
the evaluation budget; deliberate re-reads use a new draw index and are
charged. Table~\ref{tab:cost} reports the budget spent by each of the 17 documented runs. Optimizer calls range from 16 to 45. Cache hit rates are 60--69\% except for NDR with GPT-5.2 (15\%); the logs summarized here do not establish the cause of that difference.

\begin{table}[t]
\centering\small
\caption{Cost of 17 documented AtomSearch runs. Item evaluations are charged to the budget and are not necessarily individual API requests; cache hit rate is the share of
evaluation requests served from the cache. LiveMath: LiveMathematicianBench.}
\label{tab:cost}
\vspace{4pt}
\begin{tabular}{@{}cccccc@{}}
\toprule
Target & Benchmark & Attempts & Item evals. & Optimizer calls & Cache hit rate \\
\midrule
deepseek-v4-flash & NDR              & 20 & 4{,}345  & 16 & 65\% \\
                  & SpreadsheetBench & 30 & 3{,}360  & 29 & 64\% \\
                  & SearchQA         & 20 & 10{,}600 & 19 & 64\% \\
                  & LiveMath         & 20 & 923      & 19 & 65\% \\
\midrule
GPT-5.5           & NDR              & 20 & 6{,}865  & 20 & 60\% \\
                  & SearchQA         & 30 & 17{,}200 & 35 & 65\% \\
                  & LiveMath         & 30 & 1{,}474  & 27 & 64\% \\
\midrule
GPT-5.2           & NDR              & 30 & 10{,}760 & 24 & 15\% \\
                  & SpreadsheetBench & 30 & 2{,}920  & 36 & 68\% \\
                  & SearchQA         & 30 & 14{,}000 & 27 & 66\% \\
                  & LiveMath         & 30 & 1{,}295  & 29 & 67\% \\
\midrule
Qwen3.5-4B        & NDR              & 20 & 4{,}755  & 23 & 68\% \\
                  & SearchQA         & 30 & 15{,}400 & 25 & 65\% \\
                  & LiveMath         & 30 & 1{,}348  & 45 & 69\% \\
\midrule
Qwen3.6-35B-A3B   & SpreadsheetBench & 30 & 3{,}520  & 23 & 62\% \\
                  & SearchQA         & 30 & 15{,}800 & 28 & 66\% \\
                  & LiveMath         & 30 & 1{,}440  & 26 & 65\% \\
\bottomrule
\end{tabular}
\end{table}

\section{Validation-to-Test Comparison Details}
\label{app:val-test}

Table~\ref{tab:val-test} reports six comparisons, comprising SpreadsheetBench with deepseek-v4-flash and NDR with five targets. This analysis includes comparisons with an executed SkillOpt baseline, excluded validation gains below five percentage points, and excluded LiveMathematicianBench, whose validation panel has 18 items. We treat these six comparisons as a descriptive subset rather than an estimate over the full task--model matrix.

Each gain is measured relative to that run's recorded initial skill score. Initial evaluation scores can differ between methods, so the ratios do not normalize away all differences in initialization, evaluation draws, or search budget. A ratio above 100\% means the recorded test gain exceeds the recorded validation gain; it does not mean that accuracy exceeds 100\%. The ratio can also be large because its denominator is small. We therefore report both absolute gains beside each ratio and keep final test accuracy as the primary result.

The reported median ratios are 123.96\% for AtomSearch and 82.09\% for SkillOpt before rounding. These values describe the recorded point estimates. Repeated evaluation of fixed checkpoints would be needed to quantify their uncertainty and separate selection effects from differences in the learned skills.

\end{document}